\pdfoutput=1
\documentclass[11pt]{article}

\usepackage{acl}

\usepackage{times}
\usepackage{latexsym}
\usepackage{amsmath}
\usepackage{amssymb}
\usepackage{booktabs}
\usepackage{tabularray}
\usepackage{sidecap}

\usepackage[T1]{fontenc}
\usepackage[utf8]{inputenc}

\usepackage{microtype}

\usepackage{inconsolata}

\usepackage{graphicx}
\graphicspath{ {./images/} }

\title{Automatic Mapping of Anatomical Landmarks from Free-Text Using Large Language Models: Insights from Llama-2}

\author{
 \textbf{Mohamad Abdi\textsuperscript{1}},
 \textbf{Gerardo Hermosillo Valadez\textsuperscript{1}},
 \textbf{Halid Ziya Yerebakan\textsuperscript{1}}
\\
 \textsuperscript{1}Siemens Healthineers, Malvern, PA, United States
\\
 \small{
   \textbf{Correspondence:} \href{mailto:arya-abdi@siemens-healthineers.com}{arya-abdi@siemens-healthineers.com}
 }
}

\begin{document}
\maketitle
\begin{abstract}

Anatomical landmarks are vital in medical imaging for navigation and anomaly detection. Modern large language models (LLMs), like Llama-2, offer promise for automating the mapping of these landmarks in free-text radiology reports to corresponding positions in image data. Recent studies propose LLMs may develop coherent representations of generative processes. Motivated by these insights, we investigated whether LLMs accurately represent the spatial positions of anatomical landmarks. Through experiments with Llama-2 models, we found that they can linearly represent anatomical landmarks in space with considerable robustness to different prompts. These results underscore the potential of LLMs to enhance the efficiency and accuracy of medical imaging workflows.

\end{abstract}

\section{Introduction}

An anatomical landmark is a biologically meaningful point on an organism that aids in image navigation and serves as critical evidence for anomaly diagnosis within the medical imaging domain \cite{zhan2016robust, zhan2008active}. In radiology reporting, these landmarks are pivotal for precisely describing specific points within the body, acting as reference points for locating abnormalities or guiding interventions. Radiologists routinely interpret free-text reports and map the described anomalies to their corresponding positions in the image data. The advent of modern large language models (LLMs) offers promising potential to automate this task, given their advanced capabilities in interpreting and synthesizing radiology reports \cite{bhayana2024chatbots, bosbach2024ability, yan2022radbert, cai2021chestxraybert}.

Despite remarkable proficiency of LLMs in handling complex tasks \cite{kim2024language, guan2023leveraging, ouyang2022training}, these models were traditionally believed to learn only large-scale correlations by predicting the next token in a sequence, without understanding the underlying generative process \cite{bender2020climbing, bisk2020experience}. However, emerging research suggests an alternative hypothesis: these models may indeed develop coherent and interpretable representations of the generative processes they are exposed to \cite{li2022emergent, nanda2023emergent, li2021implicit, patel2021mapping, abdou2021can}. This insight opens new avenues for leveraging LLMs in automating the interpretation and mapping tasks in radiology, thereby enhancing the efficiency and accuracy of medical imaging workflows.

Motivated by these remarkable capabilities and the belief in LLMs' ability to learn coherent representations, we set out to explore the following question: "Can LLMs' internal neural activations accurately represent the spatial positions of anatomical landmarks?" We conducted experiments to linearly probe the internal neural activations of Llama-2 \cite{touvron2023llama} models to predict anatomical landmark positions. Our findings reveal:

\begin{enumerate}
  \item Llama-2 models can linearly represent the anatomical landmarks in space.
  \item These representations demonstrate considerable robustness to different prompts.
  \item Llama-2 models may linearly represent the size of anatomical landmarks.
\end{enumerate}

\section{Related Works}

\subsection{Representation Engineering}

Several studies have explored whether Large Language Models (LLMs) can offer interpretable representations of learned generative processes. This line of investigation is based on representation engineering, which suggests that high-level information can be represented through neural activities, often using the most representative or final token \cite{zou2023representation}. Leveraging this framework, Marks et al. demonstrated that that language models can linearly represent the truth or falsehood of factual statements, with these representations exhibiting directionality \cite{marks2023geometry}. Similarly, Gurnee et al. employed this methodology to reveal that the family of Llama-2 models can linearly represent both space and time, unveiling the presence of specialized neurons within Llama-2 models that activate as a function of spatial and temporal dimensions \cite{gurnee2023language}. Others have demonstrated that generative pre-training language models can effectively represent the game state of chess in a linear fashion \cite{toshniwal2022chess, nanda2023emergent}.

\subsection{LLMs as world models}

Gurnee et al. proposed that Llama-2 models could be considered world models. To support this, they created three datasets of place names at different spatial scales: globally, within the United States, and within New York City. They used these place names as inputs to the Llama-2 models and applied a linear probe to the hidden state activations to predict the spatial coordinates of the places. By comparing the distance metrics from the linear probe with those from a nonlinear probe, they showed that Llama-2 models can linearly represent spatial features. Additionally, using a block holdout generalization test, they demonstrated that these linear features are directional \cite{gurnee2023language}. 

Building on these findings, our study investigates whether the Llama-2 model can decode spatial information at a smaller scale, specifically focusing on its ability to represent the spatial positions of anatomical landmarks .

\begin{figure*}
  \includegraphics[width=1\linewidth]{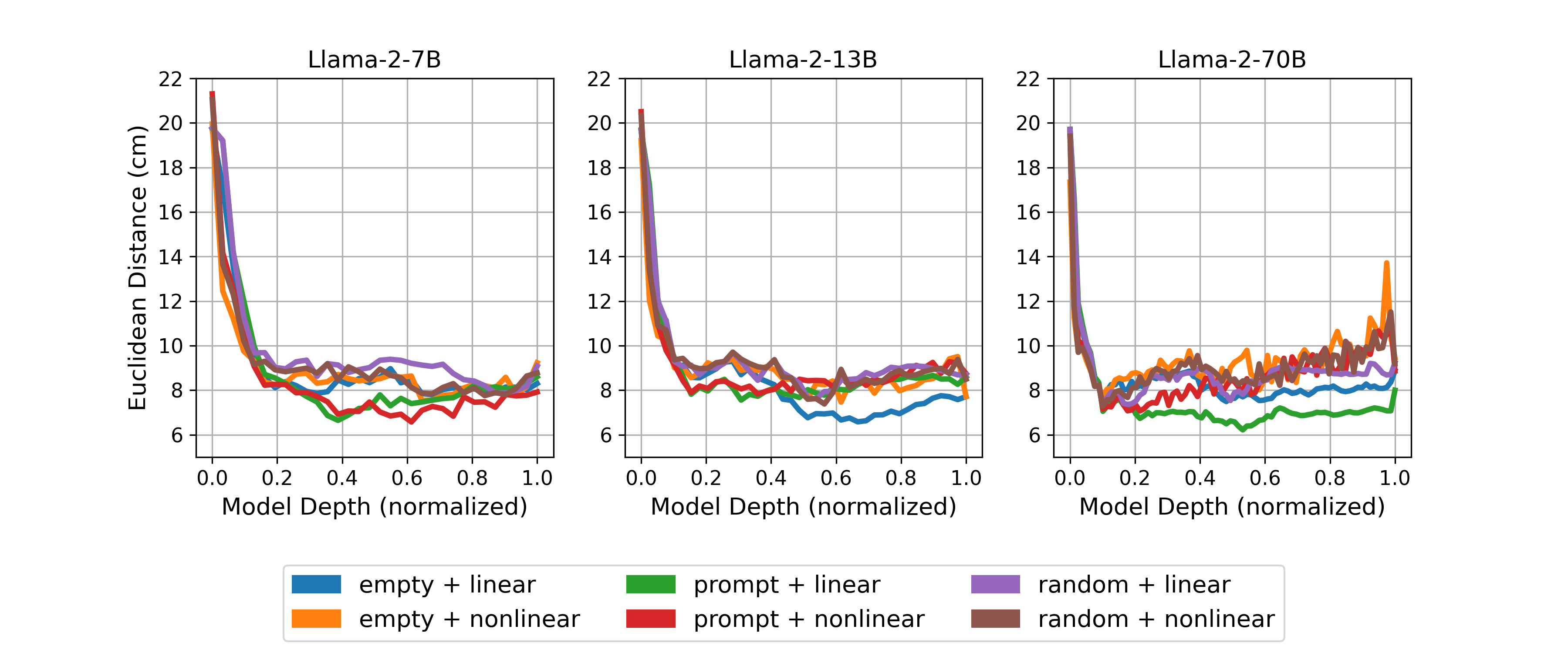}
  \caption {Mean Euclidean distance as a function of model depth for the test set. The predicted positions obtained using the linear and nonlinear (MLP) probes were compared to their corresponding values. Each probe was trained and applied to the random prompt (random), context-inducing prompt (prompt), and the unprompted (empty) activation datasets.}
    \label{fig_1}
\end{figure*}

\section{Methods}

\subsection{Datasets}

To construct our dataset, we included $n=117$ anatomical landmark names and their corresponding spatial coordinates. We utilized the Total Segmentator model \cite{wasserthal2023totalsegmentator}, an open-source segmentation toolbox \footnote{\url{https://github.com/wasserth/TotalSegmentator}}, on a human atlas to generate segmentation masks for each landmark. Each segmentation mask was then used to compute two types of spatial coordinates for each anatomical landmark in a normalized coordinate system:

\begin{enumerate}
    \item The center of the segmentation mask, used as a single-point coordinate representing the position of the anatomical landmark.
    \item The eight corners of the cube covering the segmentation mask, used to compute the coordinates of the bounding 3-dimensional box.
\end{enumerate}

We partitioned the dataset into training and test sets using a 70:30 split ratio.

\subsection{Llama-2 Encoding}

We used Llama 2.0 models \cite{touvron2023llama}, a series of pre-trained and fine-tuned auto-regressive transformers with scales ranging from 7 billion to 70 billion parameters. A commercial license was acquired prior to using the Llama 2.0 models. Each anatomical landmark name was processed through the model, sometimes with a short prompt prepended. We saved the activations of the hidden state on the last token (excluding the end-of-sequence token) for each layer. For n anatomical landmarks, this resulted in a collection of activation vectors \(z\in\mathbb{R}^m\), with m being the length of the hidden states, yielding a dataset \(x\in\mathbb{R}^{n\times m}\). These datasets were computed for each layer of the Llama 2.0 model and used in the probing experiments described in the following section.

\subsection{Probing}

We employed a probing strategy \cite{belinkov2022probing, zou2023representation} to investigate the spatial linear representation hypothesis. Specifically, we fit a ridge linear regression model to the activation dataset to predict the positions of the anatomical landmarks, either as single-point or bounding-box targets. Assuming \(y \in \mathbb{R}^{3}\) as the target values and the activation dataset \(x \in \mathbb{R}^{n \times m}\), we sought to solve the following equation:

\begin{equation} \label{eq:opt}
  \hat{\theta} = \underset{\theta}{\mathrm{arg min}} \| f_{\theta}(x) - y\|_{2}^{2} + \lambda R(x)
\end{equation}

In Equation \ref{eq:opt}, $f$, $R$, and $\theta$ denote the regression model, the regularization operator, and  the weights of the regression model respectively. For the ridge regression, the $l_{2}$ norm was used for regularization. Our analysis was based on the idea that high predictive performance on out-of-sample data indicates that the base model has spatial information that is linearly decodable in its representations \cite{gurnee2023language}. The linear regression model was implemented using the scikit-learn package. 

We compared the results of the linear probes to those of nonlinear probes. For the nonlinear probes, we used multi-layer perceptrons (MLP) with a single hidden layer, 256 neurons in the hidden layer, the Gaussian Error Linear Unit (GELU) as the activation function, and dropout with a probability factor $p=0.5$ for regularization. The nonlinear regressions were implemented using PyTorch libraries.

\begin{table*}
\centering
\renewcommand{\arraystretch}{1.2}
\begin{tabular}{@{\extracolsep{0pt}}cccccccc@{}}
\toprule
   & \multicolumn{2}{c}{\textbf{Llama-2-7B}} & \multicolumn{2}{c}{\textbf{Llama-2-13B}} & \multicolumn{2}{c}{\textbf{Llama-2-13B}} & \textbf{Baseline} \\ 
   \cline{2-3} \cline{4-5} \cline{6-7} \cline{8-8}
   & Linear & MLP & Linear & MLP & Linear & MLP & \\
   \midrule
Distance & 8.3$\pm$5.6 & 8.2$\pm$5.8 & 7.9$\pm$6.6 & 8.0$\pm$6.5 & 6.9$\pm$5.8  & 7.3$\pm$5.9  & 20.7$\pm$16.2  \\ 
DICE & 0.69$\pm$0.24 & 0.61$\pm$0.22 & 0.73$\pm$0.26 & 0.61$\pm$0.21  & 0.75$\pm$0.21  & 0.64$\pm$0.20  & 0.10$\pm$0.21 \\ [1pt]
\bottomrule
\end{tabular}
\caption{At 20\% depth of the Llama-2 models, we measured the mean and standard deviation of Euclidean distances, as well as the DICE scores for the single-point and bounding-box targets. Additionally, we included predictions from lexical similarity as the baseline. The values from the Llama-2 models are provided for both linear and nonlinear (MLP) probes.}
  \label{table_sum}
\end{table*}

\subsection{Experiments and evaluations}

\textbf{Prompting:} In their study, Gurnee et al. \cite{gurnee2023language} found little to no effect on spatial and temporal representations when prompting the Llama-2 models. Given the different spatial scale of anatomical landmarks compared to Gurnee et al.'s dataset, we employed the same prompting strategy to investigate if a similar trend could be observed. To this end, activation datasets were generated using a random prompt and a context-inducing prompt that asked for the position of the anatomical landmark. Separate linear and nonlinear probes were fitted to the random prompt, context-inducing prompt, and an unprompted training dataset. The fitted models were then applied to the test set to predict the positions of the corresponding anatomical landmarks. We used the average Euclidean distance between the predicted and target coordinates as the measure of performance.

\textbf{Spatial representation:} We further investigated whether the size of the anatomical landmarks can be decoded from the Llama-2 representations, in addition to their spatial position. To this end, we fitted linear and nonlinear regressions to the activation datasets using the bounding-box coordinates as the target values. The predicted bounding-box coordinates on the test samples were compared to the target values using the DICE score defined in Equation \ref{eq:dice}, where P and T are the volumes covered by the predicted and target bounding-boxes, respectively, and $\cap$ is the intersection operator. The DICE score measures the similarity or overlap between two sets.

\begin{equation}
    \label{eq:dice}
    DICE = \frac{2 |P \cap T|}{|P|+|T|}
\end{equation}

\textbf{Baseline:} We established a baseline for comparison of our results. For each anatomical landmark in the test set, we utilized a lexical similarity measure to identify the most similar name within the same set. The (single-point or bounding-box) coordinates of this most similar name were then used as an estimate for the coordinates of the query landmark. Specifically, we employed the Jaccard index, as defined in Equation \ref{eq:jaccard}, for two sets Q and P, to measure lexical similarity.

\begin{equation}
    \label{eq:jaccard}
    J = \frac{|Q \cap T|}{|Q \cup T|}
\end{equation}

\section{Results}

Figure \ref{fig_1} illustrates the mean Euclidean distance between the predicted and target positions of anatomical landmarks within the test set as a function of model depth. Deeper layers of the Llama-2 models yield more accurate spatial representations, with accuracy approaching a plateau at approximately 20\% depth across all Llama-2 model sizes. We did not observe a significant advantage in prompting the Llama-2 models. Furthermore, the more complex nonlinear regression did not produce significantly more accurate predictions compared to linear regression. These findings suggest that the spatial positions of anatomical landmarks can be effectively represented linearly using the Llama-2 models

Table \ref{table_sum} presents the Euclidean distances (mean ± standard deviation) at 20\% depth of the Llama-2 model for the single-point test data. An increase in model size results in more accurate predictions, a trend consistent with findings from previous study \cite{gurnee2023language, hoffmann2022training}. Notably, the predictions from the Llama-2 models are significantly more accurate than the baseline.

\begin{SCfigure}
  \includegraphics[width=.4\linewidth, height=1\linewidth]{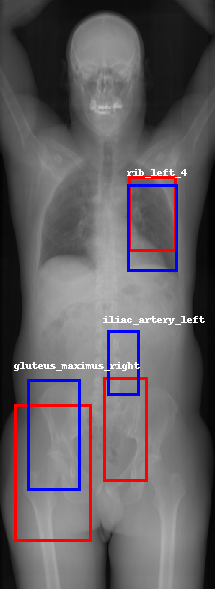}
  \caption {Example predicted (in blue) and target (in red) bounding boxes overlaid on a human atlas. Predicted bounding boxes were computed by applying the linear probe to the activation layers at 20\% depth of Llama-2 70B model.}
  \label{fig_bbox}
\end{SCfigure}

Table \ref{table_sum} also provides the Dice scores (mean ± standard deviation) at 20\% depth of the Llama-2 model for the bounding-box test data. Similar trends were observed, with larger models yielding better predictions of both the size and position of the anatomical landmarks. However, the Dice scores from the linear probe are slightly better than those from the nonlinear probe. There is also a significant difference between the baseline and the Llama-2 models. Examples of the predicted bounding boxes (blue) and target bounding boxes (red) are visualized in the coronal view in Figure \ref{fig_bbox}. These results suggest that the Llama-2 model may represent the spatial size of anatomical landmarks in addition to their position, although further ablation studies are needed to substantiate this finding.

\section{Conclusion}

Our study demonstrates the potential of modern large language models, particularly Llama-2, for representing the position of the anatomical landmarks from free-text input data. While a deeper understanding of the mechanisms behind these representations and their extent is needed, our findings hint at the potential of these models to alleviate workload in radiology workflows. Future research can delve into deeper into this specific use-case LLMs in radiology, utilizing a broader range of anatomical landmarks and LLMs tailored specifically for medical applications.

\section{Limitations}

One limitation of this methodology is the limited number of landmarks used in both training and testing. A broader set could offer stronger evidence of LLMs' spatial representation. Additionally, the method underutilizes LLMs' few-shot learning capabilities. An alternative could integrate known landmark coordinates into the prompt, providing context, and query for unknown landmark coordinates.

\bibliography{main}

\end{document}